\documentclass[journal]{IEEEtran}

\usepackage{amsmath}
\usepackage{amssymb}
\usepackage{subfigure}
\usepackage{graphicx}
\usepackage{url}
\usepackage{cite}
\usepackage{multirow}
\usepackage{makecell}
\usepackage{booktabs}
\usepackage{bm}
\usepackage{algorithm}
\usepackage{algorithmic}

\begin{document}

%\tableofcontents

\title{Detection of Deepfake Videos Using Long Distance Attention}

\author{Wei Lu, \IEEEmembership{Member,~IEEE},
        Lingyi Liu,
        Junwei Luo,
        Xianfeng Zhao, \IEEEmembership{Senior~Member,~IEEE},\\
        Yicong Zhou, \IEEEmembership{Senior~Member,~IEEE},
        Jiwu Huang, \IEEEmembership{Fellow,~IEEE},

%\thanks{This work is supported by the National Natural Science Foundation of China (No. U2001202, No. 62072480, No. U19B2022, No. U1736118), the National Key R\&D Program of China (No. 2019QY2202, No. 2019QY(Y)0207), the Key Areas R\&D Program of Guangdong (No. 2019B010136002, No. 2019B010139003), the Key Scientific Research Program of Guangzhou (No. 201804020068), Shenzhen R\&D Program (No. GJHZ20180928155814437).}
\thanks{Wei Lu, Lingyi Liu and Junwei Luo are with the School of Computer Science and Engineering, Guangdong Province Key Laboratory of Information Security Technology, Ministry of Education Key Laboratory of Machine Intelligence and Advanced Computing, Sun Yat-sen University, Guangzhou 510006, China
(e-mail: luwei3@mail.sysu.edu.cn, liuly83@mail2.sysu.edu.cn, luojw8@mail2.sysu.edu.cn)}
\thanks{Xianfeng Zhao is with the State Key Laboratory of Information Security, Institute of Information Engineering, Chinese Academy of Sciences, Beijing 100195, China, and also with the School of Cyber Security, University of Chinese Academy of Sciences, Beijing 100195, China (e-mail: zhaoxianfeng@iie.ac.cn)}
\thanks{Yicong Zhou is with the Department of Computer and Information Science, University of Macau, Macau 999078, China (e-mail: yicongzhou@um.edu.mo).}
\thanks{Jiwu Huang is with the Guangdong Key Laboratory of Intelligent Information Processing and Shenzhen Key Laboratory of Media Security, Shenzhen University, Shenzhen 518060, China, and also with the Shenzhen Institute of Artificial Intelligence and Robotics for Society, Shenzhen 518055, China (e-mail: jwhuang@szu.edu.cn).}
}

\maketitle

\begin{abstract}
With the rapid progress of deepfake techniques in recent years, facial video forgery can generate highly deceptive video contents and bring severe security threats.
And detection of such forgery videos is much more urgent and challenging.
Most existing detection methods treat the problem as a vanilla binary classification problem.
In this paper, the problem is treated as a special fine-grained classification problem since the differences between fake and real faces are very subtle.
It is observed that most existing face forgery methods left	 some common artifacts in the spatial domain and time domain, including generative defects in the spatial domain and inter-frame inconsistencies in the time domain.
And a spatial-temporal model is proposed which has two components for capturing spatial and temporal forgery traces in global perspective respectively.
The two components are designed using a novel long distance attention mechanism.
The one component of the spatial domain is used to capture artifacts in a single frame, and the other component of the time domain is used to capture artifacts in consecutive frames.
They generate attention maps in the form of patches.
The attention method has a broader vision which contributes to better assembling global information and extracting local statistic information.
Finally, the attention maps are used to guide the network to focus on pivotal parts of the face, just like other fine-grained classification methods.
The experimental results on different public datasets demonstrate that the proposed method achieves the state-of-the-art performance,
and the proposed long distance attention  method can effectively capture pivotal parts for face forgery.
\end{abstract}


\section{Introduction}
\label{sec:intro}

The deepfake videos are designed to replace the face of one person with another's.
The advancement of generative models \cite{gan,vae,progressivegan,TNNLSlookmore} makes deepfake videos become very realistic.
In the meantime, the emergence of some face forgery applications \cite{deepfake,fakeapp,faceswap} enables everyone to produce highly deceptive forged videos.
Now, the deepfake videos are flooding the Internet.
In the internet era, such technology can be easily used to spread rumors and hatred, which brings great harm to society.
Thus the high quality deepfake videos that cannot be distinguished by human eyes directly have aroused interest among researchers.
An effective detection method is urgently needed.

\begin{figure}
	\centering
	{\includegraphics[width=0.46\textwidth]{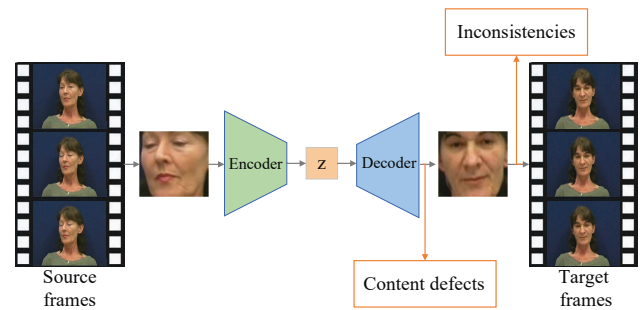}}
	\caption{The generation process of deepfake videos. The original video is divided into frames and cropped out of the faces.
		The target face is generated by an encoder-decoder which introduces content defects.
		Then the target face is spliced back to the original frame, and  inconsistencies are introduced.
		Finally, all the frames are synthesized into a fake video.}
	\label{fig:forgery_flow}
\end{figure}

The general process of generating deepfake videos is shown in Fig. \ref{fig:forgery_flow}.
Firstly, the video is divided into frames and the face in each frame is located and cropped.
Then, the original face is converted into the target face by using a generative model and spliced into the corresponding frame.
Finally, all frames are serialized  to compose the deepfake video.
In these processes, two kinds of defects are inevitably introduced.
In the process of generating forged faces, the visual artifacts in the spatial domain are introduced by the imperfect generation model.
In the process of combining frame sequences into videos, the inconsistencies between frames are caused by the lack of global constraints.

Many detection methods are proposed \cite{mediaforensics161,mesonet,landmark} based on the defects in the spatial domain.
Some of the methods take advantage of the defects of face semantics in deepfake videos,
because the generative models lack global constraints in the process of fake face generation,
which introduces some abnormal face parts and mismatched details in the face from a global perspective.
For example, face parts with abnormal positions \cite{landmark}, asymmetric faces \cite{asymmetry}, and eyes with different colors \cite{mediaforensics161}.
However, it's fragile to rely entirely on these semantics.
Once the deepfake videos do not contain the specific semantic defects that the method depends on,
the performance will be significantly degraded.

There are also some ``deep'' approaches \cite{mesonet,recurrent,twostream}, which attempt to excavate spatial defects according to the characteristics of the deepfake generators.
However, compared with image contents, the forgery traces in the spatial domain are very weak, and the convolutional networks tend to extract image content features rather than the traces \cite{bayar}.
So blindly utilizing deep learning is not very effective in catching fake contents \cite{fakecatcher}.

Since the deepfake video is synthesized frame by frame, and there is no precise constraint between the frame sequences,
the inconsistencies in the time domain will be introduced.
Some methods exploit these defects of the time domain.
The movements of eyes are exploited in \cite{eyemovement}.
Li et al. \cite{eyeblinking} use the human blink frequency in the video to detect the deepfake videos.
The movement of lip \cite{lipmovement} and the heart rate \cite{heartrate} are also exploited as the identification basis between authentic videos and deepfake videos in the time domain.
The optical flows and the movement patterns of the real face and fake face are classified in \cite{optical} and \cite{motionpattern}, respectively.

All of the methods mentioned above take the deepfake detection as a vanilla binary classification problem.
However, as the counterfeits become more and more realistic, the differences between real and fake ones will become more and more subtle and local which making such global feature-based vanilla solutions work not well \cite{multiattentional}.

Similar problems have been studied in the field of fine-grained classification.
Fine-grained classification aims to classify very similar categories, such as species of the bird, models of the car, and types of the aircraft \cite{seebetterbefore}.
Since the deepfake detection and fine-grained classification share the same spirit, that learning subtle and discriminative features,
in \cite{multiattentional}, the deepfake detection is reformulated as a fine-grained classification task.
And a convolutional attention module with $1 \times 1$ is adopted to make a network focus on the subtle but critical regions.

However, combining global semantics is just as important as focusing on local areas.
Because some defects are normal from a local or isolated perspective, but abnormal from a global perspective.
For example,  uncoordinated head postures \cite{headpose}, mismatched facial expressions and head movements \cite{expression}, and mismatched eye details  \cite{mediaforensics}.
These kinds of defects exist between different parts of the face at a long distance.
In other words, the local areas of focus should be determined according to the global semantics \cite{TNNLSglobal}, and modeling long distance dependencies in both spatial domain and time domain is important.
But it is not directly for the convolutional attention mechanism, especially when the kernel is small.
The global pooling may be a choice to assembling global information, however, the weak forgery clues will be averaged by this operation, and resulting in a loss of distinguishability  \cite{multiattentional}.

\begin{figure}
	\centering
	{\includegraphics[width=0.41\textwidth]{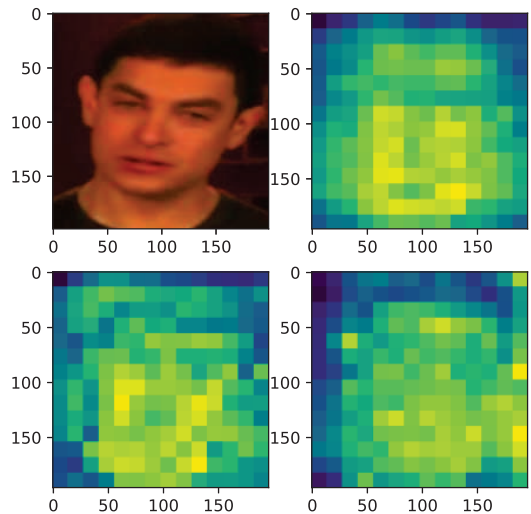}}
	\caption{The attention maps are generated by the novel long distance attention  mechanism. Pivotal facial regions are emphasized in patches by these maps.}
	\label{fig:attention_intro}
\end{figure}

Vision Transformer (ViT) \cite{ViT} is a widely used model, which can draw global dependencies and assemble global information relying entirely on a self-attention mechanism.
However, according to our experiments, as well as some existing works \cite{ConvoViT} \cite{ViTDistill},
the effect of applying ViT directly to the deepfake detection task is general.
Thus, we draw lessons from the fine-grained classification and propose a novel long distance attention mechanism according to the characteristics of deepfake videos.
The long distance attention mechanism is designed to determine the pivotal parts of forgery by assembling information from a global perspective.
We adopt the long distance attention in our spatial-temporal model to exploit the defects in the spatial domain and time domain.
The spatial-temporal model is used to generate attention maps in the form of patches and guides the network to focus on pivotal local parts of the face.
An example of our attention maps is shown in Fig. \ref{fig:attention_intro}, the pivotal parts of the face are emphasized as highlights.

The contributions of this paper are summarized as follows:
\begin{itemize}
		\item
		The experience of the fine-grained classification field is introduced, and a novel long distance attention mechanism is proposed which can generate guidance by assembling global information.
        % A novel long distance attention mechanism is proposed and  used  to construct a multi-level semantic guidance model based on spatial defects in single frame and temporal defects in multi frame.
    	\label{step:con1}
		\item
        It confirms that the attention mechanism with a longer attention span is more effective for assembling global information and highlighting local regions.
        And in the process of generating attention maps, the non-convolution module is also feasible.
        \label{step:con2}
        \item
        % We have done extensive experiments to demonstrate the effectiveness of out attention method, and the model outperforms other state-of-the-art detection methods on mainstream deepfake datasets.
         %The experience of fine-grained classification is introduced and referenced to design a spatial-temporal attention model according to the characteristics of deepfake, the experimental results show that it achieves the state-of-the-art performance.
         A spatial-temporal model is proposed to capture the defects in the spatial domain and time domain,  according to the characteristics of deepfake videos, the model adopts the long distance attention as the main mechanism to construct a multi-level semantic guidance. The experimental results show that it achieves the state-of-the-art performance.
        \label{step:con3}
\end{itemize}

The remainder of this paper is organized as follows.
In Section \ref{sec:related work}, we first discuss the related work in the field of fine-grained classification.
Then, the classical Vision Transformer is introduced briefly.
In Section \ref{sec:forgeryanalysis},
we analyze the defect characteristics of deepfake videos.
In Section \ref{sec:Approach}, the proposed method is introduced in details.
Section \ref{sec:experiments} discusses the experimental results. The ablation analysis is given in Section \ref{sec:ablationanalysis}. The conclusion is presented in Section \ref{sec:conclusion}.

\section{Related works}
\label{sec:related work}

\subsection{Fine-grained classification}
\label{sec:fine-grained}

In the past few years, the performance of general image classification tasks has been significantly improved.
From the amazing start of Alexnet \cite{AlexNet} in Imagenet \cite{imagenet}, the method based on deep learning almost dominate the Imagenet competition.
However, for fine-grained object recognition \cite{weaklysupervised,TNNLSdisc,Bilinearcnn,twolevel,TNNLSsimple},
there are still great challenges.
The main reason is that the two objects are almost the same from the global and apparent point of visual.
Therefore, how to recognize the subtle differences in some key parts is a central theme for fine-grained recognition.
Earlier works \cite{partstacked,SPDACNN} leverage human-annotated bounding box of key parts and achieve good results.
But the disadvantage is that it needs expensive manual annotation,
and the location of manual annotation is not always the best distinguishing area \cite{learningmulti,viewdependent},
which completely depends on the cognitive level of the annotator.

Since the key step of fine-grained classification is focusing on more discriminative local areas \cite{maskcnn},
many weakly supervised learning methods \cite{seebetterbefore,learningmulti,lookcloser} have been proposed.
Most of them use kinds of convolutional attention mechanisms to find the pivotal parts for detection.
Fu et al. \cite{lookcloser} use a recurrent attention convolutional neural network (RA-CNN) to learn discriminative region attention.
Hu et al. \cite{senet} propose a channel-wise attention method to model interdependencies between channels.
In \cite{learningmulti}, a multi-attention convolutional neural network is adopted and more fine-grained features can be learned.
Hu et al. \cite{seebetterbefore} propose a weakly supervised data augmentation network using attention cropping and attention dropping.

Deepfake detection and fine-grained classification are similar, that attempt to classify very similar things.
Thus we learn from the experience in this field and leverage the attention maps generated with long range information to make the networks focus on pivotal regions.

\subsection{Vision transformer}
\label{sec:ViT}

Transformer \cite{transformer}, a kind of self-attention architectures, is initially applied in natural language processing (NLP) and shows excellent performance.
Its variant in the field of computer vision, Vision Transformer (ViT) \cite{ViT}, is first proposed by the Google team in 2020 and attracts a lot of attention.
In vision, attention is usually used as a component of convolutional networks while keeping the overall structure.
ViT shows that reliance on CNNs is not necessary.
To apply the transformer to images directly, they firstly split the image into patches and project the patches to linear embedding.
As a classification model, it generates a final discriminant vector through several stacking layers of self-attention modules.
The self-attention modules are used to integrate the features of each patch with the self-attention mechanism.
The self-attention mechanism is a stunning mechanism, which draws global dependencies and assembles global information.

It may be a promising candidate to deal with the detection of deepfake videos since the deepfake videos need to be considered from a global perspective and focused on the critical regions.
However, we find that it is not effective to apply ViT to deepfake detection directly.
Therefore, we learn from ViT and propose a novel long distance attention mechanism.
It is used to guide the backbone network to focus on critical regions by assembling global information.

\begin{figure}
        \centering
             \subfigure[Local defects]{\includegraphics[width=0.4\textwidth]{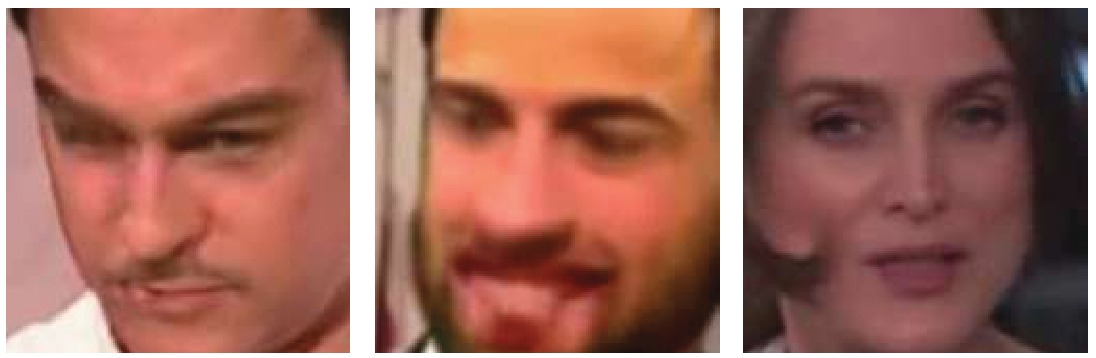}}
             \subfigure[Mismatched eyes]{\includegraphics[width=0.4\textwidth]{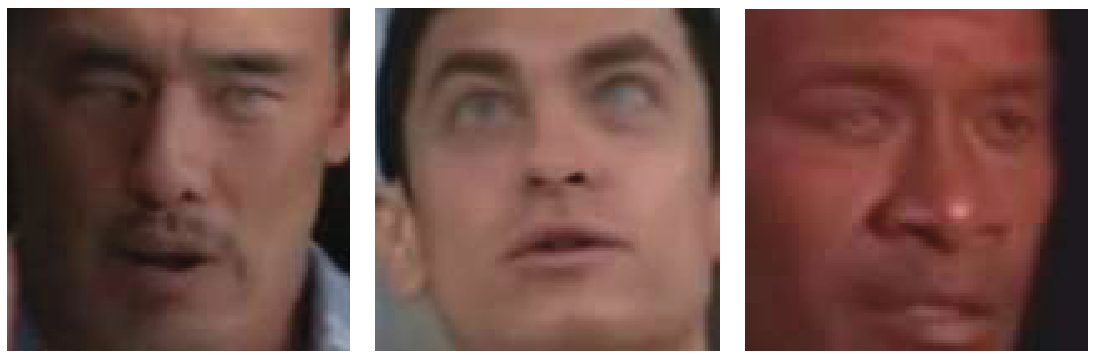}}
             \subfigure[Abnormal structures]{\includegraphics[width=0.4\textwidth]{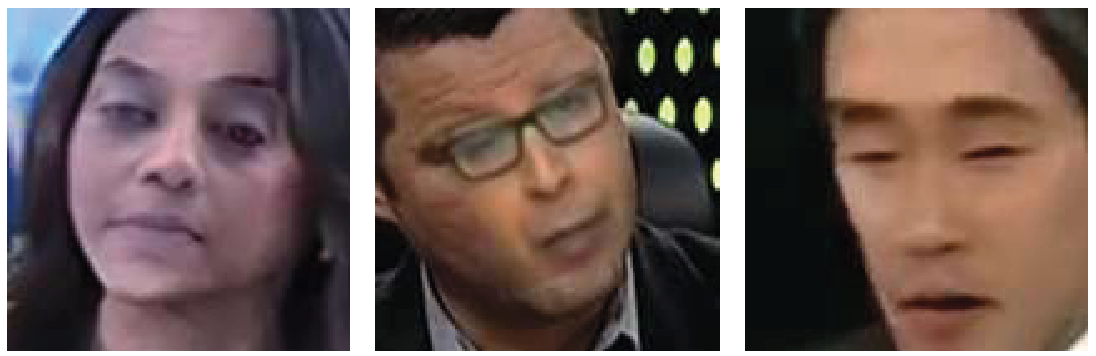}}
           \caption{Some typical defects of deepfake videos in the spatial domain. The images in the first row reflect some local defects, i.e.,
           obvious forgery clues in the mouth of the left and middle pictures, and a strange facula near the hair in the right picture.
           The second row contains faces with weird eyes. The third row contains faces with abnormal face structure.
           }
        \label{fig:fakesample}
\end{figure}

\begin{figure*}
        \centering
             {\includegraphics[width=0.9\textwidth]{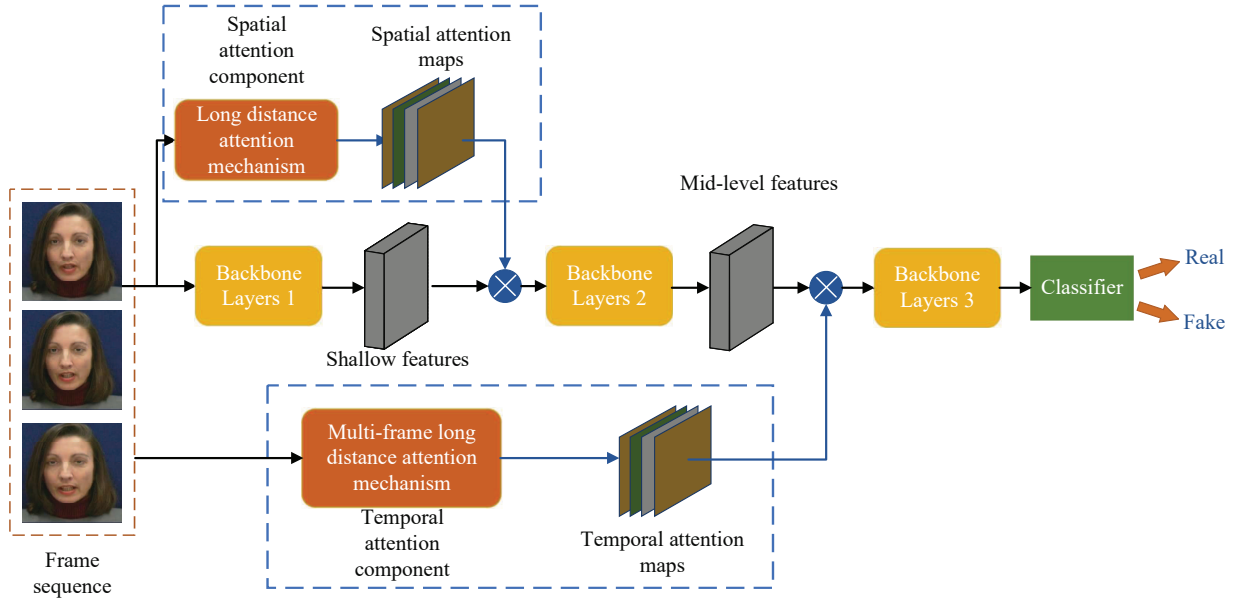}}
         \caption{The framework of the proposed method. There are two essential components in the framework.
         A spatial attention module for capturing the spatial defects in a single frame, and a temporal attention module for capturing the temporal inconsistencies between consecutive frames.
         The components are used to generate guidance to make the backbone network focus on pivotal local regions.}
        \label{fig:frame_work}
\end{figure*}

\section{Analysis of deepfake}
\label{sec:forgeryanalysis}

The deepfake videos, generated by GANs \cite{gan} and VAEs \cite{vae}, are formidably realistic and difficult for human eyes to discriminate.

Since the differences between authentic videos and deepfake videos are subtle,
detectors that blindly utilizing deep learning are not effective in catching fake content \cite{fakecatcher}.
Similar problems have been studied in the field of fine-grained classification.
A crucial experience is that using an attention mechanism to make the network focus on pivotal local regions can greatly improve the classification performance.

The generative models also have some inherent defects, which make deepfake detection possible.
Whether it's  GANs or VAEs,
the generative networks will have an up-sampling process in the generation process to generate high-resolution images from latent coding \cite{gan,vae}.
This allows the network to fill in details into the rough image.
Deconvolution allows the model to draw a larger square from a point in the small graph.
However, deconvolution is prone to uneven overlap, especially when the kernel size cannot be divided by the step size.
In theory, the neural network can learn the weight parameters carefully to avoid this kind of defect, but in fact, the neural network cannot completely avoid this kind of defect \cite{convoguide}.
This overlapping style is reflected in two dimensions. The uneven overlapping multiplication of two coordinate axes results in the image block similar to chessboard \cite{convoguide},
and resulting in a loss of facial texture details.
Liu et al. \cite{spatialphase} observe that the up-sampling is a necessary step of most face forgery techniques and utilize phase spectrum to capture the up-sampling defects of face forgery.
Since the up-sampling occurs between adjacent pixels, it is advantageous to capture the local information and collect statistics by using small blocks of appropriate size \cite{TNNLSrobust}.
On the other hand, deepfake often generates abnormal face semantics.
For example, unconvincing specular reflections in the eyes, either missing or represented as white
blobs, or roughly modeled teeth, which appear as a single white blob \cite{mediaforensics}.
The semantics and textures of the human face also appear in the form of the region \cite{TNNLSmultirepre}.
Therefore,  the processing of facial features in the form of patches is conducive to extracting local statistical information and capturing forgery traces.
In the long distance attention, the input image is divided into many non-overlapping small patches to collect local information.

However, some face semantics are normal from the local perspective but abnormal from the global perspective.
That's because the GANs lack global constraints which introduce abnormal facial parts and mismatched details.
It is observed that the density distributions of normalized face landmark locations on real and GAN-synthesized fake faces are different \cite{landmark},
because there is no coordination mechanism in the generation process of face components.
This also leads to the asymmetry of the face \cite{asymmetry}.
In addition to the global structure as clues, the difference of details between facial components is also a key to the detection.
For example,  human eyes are always separated by a certain distance and have the same color, but the eyes of the fake face sometimes show different color\cite{mediaforensics}.
An example of defects in the spatial domain is shown in Fig. \ref{fig:fakesample}. The first row reflects defects in a local region, and the next two rows reflect defects from a wider vision.
It is also observed that biological signals are not coherently preserved in different synthetic facial parts \cite{fakecatcher}.
Therefore, assembling global semantic information and considering the location relationship between facial components will help to find these generative defects.

In addition to the generative defects in the spatial domain, temporal defects are also existed in deepfake videos.
In \cite{eyeblinking}, the temporal inconsistencies are caught by the frequency of eye blinking.
The inconsistency is also reflected in the face motion.
The face motion patterns of real videos and deepfake videos have some differences, and can be used for classification \cite{motionpattern}.
Furthermore, there is a strong correlation between facial expression and head movement \cite{expression}. Changing the former without modifying the latter may expose a manipulation.
It is also observed that temporal consistencies of human biological signals are not well preserved in GAN-erated content \cite{fakecatcher}.
Thus, it is beneficial to modeling the continuity of face in the videos for deepfake detection.
We exploit these inconsistencies in the time domain with a temporal model.

\section{The proposed method}
\label{sec:Approach}

\begin{figure*}
        \centering
             {\includegraphics[width=0.9\textwidth]{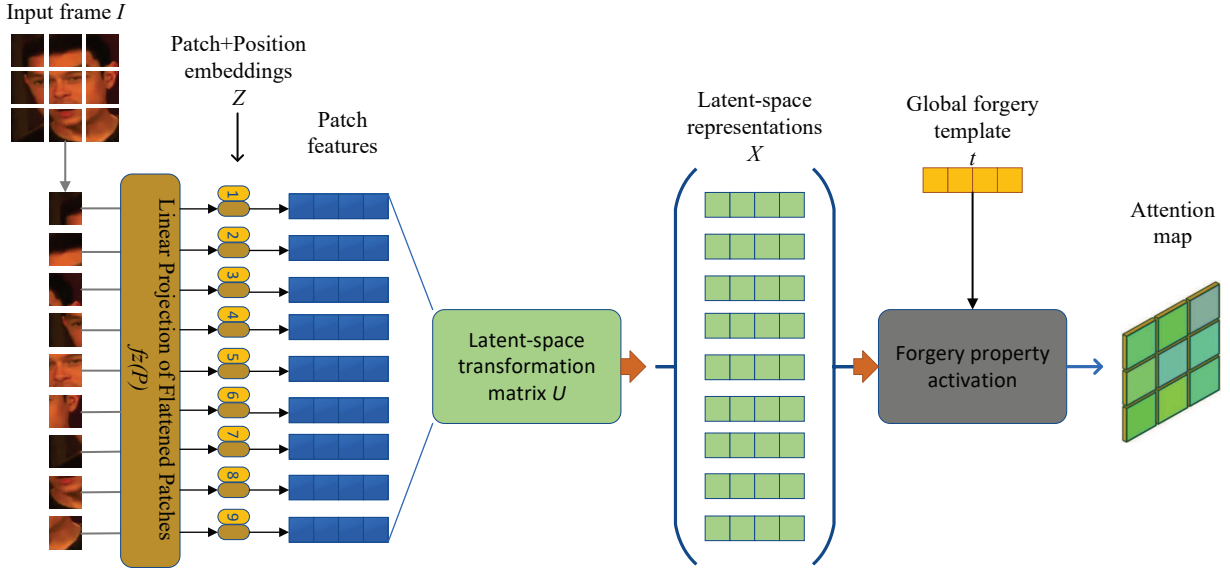}}
         \caption{The proposed long distance attention mechanism. The image is split into small patches.
         The patches are linearly projected to patch embedding and the position embedding is added. Then, the embedding is transformed to representations in a latent space by a matrix.
         Lastly, a global forgery template rectified by learning is used to activate the forgery property of each representation to generate attention maps.}
        \label{fig:attention}
\end{figure*}

\subsection{Overview}

In this section, the motivation to use long distance attention is given first and then the proposed model is described briefly.

As aforementioned, there is no precise global constraint in the deepfake generation model, which always introduces disharmony between local regions in the face forgery from a global perspective.
In addition to the artifacts that exist in each forgery frame itself, there are also inconsistencies (e.g., unsmooth lip movement) between frame sequences because  the deepfake videos are generated frame by frame.
To capture these defects, a spatial-temporal model is proposed, which has two components for capturing spatial and temporal defects respectively.
Each component has a novel long distance attention mechanism which can be used to assembling the global information to highlight local regions.
% However, the ordinary convolutional networks tend to extract the features of image content rather than those artifacts that hidden in the subtle part of face. So, to combine the overall
% To capture this disharmony and make network focusing on the pivotal parts,  a attention method is proposed which can be used to combine the overall information to highlight key parts.
% Since the long distance attention method we proposed has no limitation of length, it can be applied to multi-frame directly to capture the discontinuity.

Based on the observation \cite{dogans} that the artifacts caused by generation model mainly preserved in textural information of shallow features,
the attention maps generated by the spatial component are adopted to recalibrate the shallow features maps which are generated by the first several convolutional layers.
As the inconsistency occurs in relative high-level semantic features, the attention maps generated by the temporal attention component are used to guide the relative high-level semantic features.

The framework of the spatial-temporal model is shown in Fig. \ref{fig:frame_work}. Two essential components are integrated into the backbone network:
1) a spatial attention component for capturing spatial disharmony and focusing on shallow features.
2) a temporal attention component for capturing temporal inconsistencies and focusing on mid-level features.
Both the attention maps are used to recalibrating the feature maps and make the network focusing on pivotal local regions.
The backbone adopted is the Xception \cite{xception} which performs well in the vision field.
% and we also explore how a slight backbone influence the performance

% we focus instead on the channel relationship and propose a novel architectural unit, which we term the "Squeeze-and-Excitation"(SE) block, that adaptively recalibrates channel-wise feature^^^^
% responses by precisely modelling interdependencies between channels^^^^

% a central theme

\subsection{Long distance attention}

In faces, a semantic region is a small area with rich information, like human eyes.
Based on the observation we have mentioned, the long distance attention mechanism is proposed to model the interdependencies between the semantic regions to perform feature recalibration mediately.
It contributes to using global information to selectively emphasize informative regions and suppress regions that useless for forgery detection.

As the key parts of face forgery can be regarded  as many small areas with abnormal clues, the image is divided into many non-overlapping small patches.
These patches contain the local statistical information that might imply potential forgery clues.
And then the weight of fake confidence for a small area corresponding to each patch is obtained which is achieved by the long distance attention.

Denote the input image as $I \in \mathbb{R}^{H \times W \times C}$, and the resolution is $H \times W$, $C$ is the number of channels.
The image is divided into a sequence of small patches $P=[p_{1},p_{2},\dots,p_{N}]$. Therefore, there will be $N=HW/s$ patches, each one has $C$ channels  and the resolution $s \times s$.
Then each patch is flattened and mapped to a $D$  dimension vector with a trainable linear projection $f_z(P)$, which transfer patches to embedding $Z=[z_{1},z_{2},\dots,z_{N}]$ for ease of processing \cite{TNNLSgenera}.
Considering that the position of each patch reflects the spatial relationship between them, in order to reserve positional information, position embedding is added to the patch embedding to compose patch features \cite{lbp}.
The position embedding is shaped in a learning way.
To model the internal relationship between the patch features, a necessary global forgery template $t$ is utilized \cite{TNNLSglobal}.
The template $t$ is used to model the global association of a latent forgery property space.
In order to intuitively understand the so-called latent forgery property space, an inaccurate example is the optical flow space of the patches, the optical flow sometimes reflects an irregular variation of the deepfake videos.
Since there may be more than one forgery property space, multiple templates are adopted.
At the same time, the patch features will be mapped to the representations $X=[x_{1},x_{2},\dots,x_{N}]$ in each latent space, which is implemented with a learnable transformation matrix $U$.
Both the matrix and template are shaped in a learning way.
After that, the template in each latent forgery property space is used to consult each representation to get the forgery property activation \cite{TNNLSlatent}.
The activation is treated as the attention weight, and adopted to guide the feature maps.

As shown in Fig. \ref{fig:attention} the long distance attention consists of three main steps:
1) The patches are flattened to patch embedding $Z$ and added the position embedding to compose the patch features.
2) The patch features are mapped to the representations $X=[x_{1},x_{2},\dots,x_{N}]$ of a latent forgery property space, by a learnable transformation matrix $U$.
3) Finally, the global forgery template $t$ is used to consult each representation to obtain the activation rate of each representation.

Since the activation rates represent the confidence level of each patch with the suspicious region, they are reshaped to a attention map with the same resolution as the feature maps of the backbone,
and applied by element-wise multiplication to emphasize pivotal regions.

As aforementioned, there are not only one forgery property space, in fact, we adopt $12$ such attention module to produce different attention maps of different latent space,
and linearly combined into $m$ final attention maps for a robust and efficient reason \cite{multiattentional}, more discussion is given in \ref{sec:quantitymaps}.

\subsection{Spatial attention model}

In this section, we introduce the overall spatial attention model in details.
The spatial attention model is designed to capture the artifacts that existed in the spatial domain with a single frame.
As aforementioned, since there is no precise global constraint between face parts which will introduce disharmonious facial structures and mismatched texture details \cite{mediaforensics},
it is beneficial to generate guidance from a global aspect.
Most of the existing methods use pooling to deal with the problem \cite{senet}, such as global pooling, channel pooling and so on.
However, compared with the local association, the global association information is very weak and difficult to be established \cite{fakecatcher}.
On the other hand, defects such as oversampling and insufficient texture appear in the local area,
so an appropriate size of the local receptive field is benefiting for the collection of this statistical information.
With the long distance attention mechanism, these problems can be well balanced.

As we want to use the long distance attention mechanism to capture the defects of the spatial domain in a global perspective, a single frame of the tested video is used as the input.
And to recalibrate the importance between regions, the attention maps generated by a single frame are adopted to the feature maps of the backbone network.
As textural features exist in shallow features \cite{multiattentional}, we make the attention works with the first several layers of the backbone.
%In the implementation, the attention maps are used to guide the layer before the first block of Xception \cite{xception}.
More specifically, the input image  $I$  which is used for the backbone and the spatial attention model is reshaped to the resolution $398 \times 398$ and $224 \times 224$ respectively.
Then the convolutional feature maps are extracted by the first several layers of the backbone.
And the spatial attention module receives the relatively small image which is tackled by the attention mechanism we have described above.
Finally, the attention maps generated by the mechanism are element-wise multiplied shallow feature maps to get the emphasized feature maps.

\begin{figure}
	\centering
	{\includegraphics[width=0.48\textwidth]{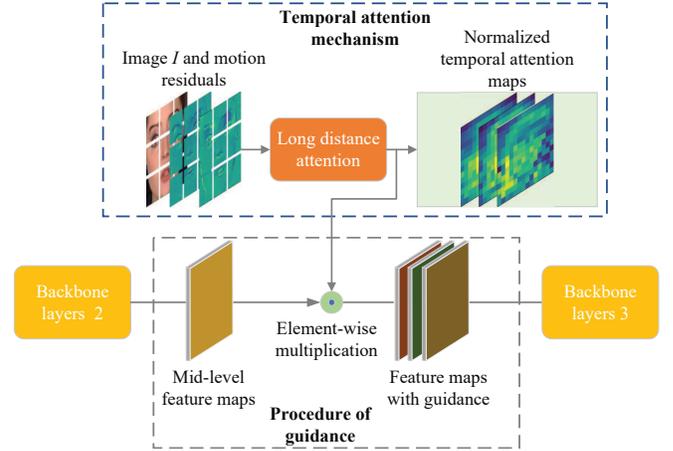}}
	\caption{The process of temporal attention generation. The image of the tested video and its following motion residuals are used as inputs.
		Then the attention maps are generated by the long distance attention and adopted to guide the mid-level feature maps.}
	\label{fig:temporal_attention}
\end{figure}

\subsection{Temporal attention model}

The movement of human faces is a complex and delicate process.
For example, the facial expressions and head movements are strongly correlated \cite{TNNLScolla} and changing the former without modifying the latter may expose a manipulation \cite{expression}.
However, since the deepfake videos are synthesized frame by frame and do not precisely model the correlation between frames, it almost inevitably  introduces inconsistencies.
In order to capture these temporal inconsistencies,  consecutive frames of video are required.
For the frame to be detected, the next $n$ frames are also utilized in the temporal attention model. The number $n$ is determined by experiments in Section \ref{sec:quantityofframes}.
For the temporal attention, we care more about the variation of videos in the time dimension, so that we calculate the motion residuals between adjacent frames as the inputs.
As shown in Fig. \ref{fig:temporal_attention}, the frame $I$ and the motion residuals are split into patches, all of the patches are composed of a sequence to be the input of the model.
%Because the positional information is formed in a learning way, it is not necessary  to determine the order of these patches in the sequence.
In this way, the template $t$ of the temporal attention model is used to model the inconsistency between frames and to obtain the activation rate of each region in a latent inconsistency space.
The activation rates are used as the attention weights, and represent the confidence of inconsistency in each region.
Since the inconsistency in the time domain is a relative high-level semantics compared with the features in the spatial domain, the temporal attention maps are applied to relative high-level feature maps.
In the same way, the attention maps are reshaped to the same size with relative high-level feature maps and element-wise multiplied.

\section{Experiments}
\label{sec:experiments}

In this section, the experiment setups are introduced firstly and then we present extensive experimental results to demonstrate the superiority of our method.

\subsection{Datasets and implementation details}
\label{subsec:data}

Two mainstream deepfake datasets are used in our experiment, including FaceForensics++ (FF++) \cite{ff++} and Celeb-DF \cite{celebDF}.
FaceForensics++ and Celeb-DF are both large-scale datasets which are widely used in face forgery detection.
FaceForensics++ dataset consists of four kinds of face forgery videos, which is generated by four state-of-the-art methods, i.e., DeepFake (DF) \cite{deepfake}, FaceSwap (FS) \cite{faceswap}, Face2Face (F2F) \cite{face2face} and NeuralTexture (NT) \cite{neuraltextures}.
For each video of FF++, it has two compression versions (i.e., HQ, LQ), which are compressed by H.264 \cite{h264} with constant rate quantization parameters set by 23 and 40.
Celeb-DF is a great challenge to the current detection methods. It consists of more than 5000 deepfake videos, and the real videos are gathered from social media.
Benefiting from an elaborate generation model, the generated videos are very realistic.
For all video frames, we use Dlib \cite{dlib} to detect and crop faces. The aligned facial images are resized to $398 \times 398$ for the backbone network and $224 \times 224$ for attention modules respectively.
And the size of all patches is set to $16\times16$.

\begin{table}
	\centering
	\caption{The quantitative comparisons among recent methods and the proposed on FaceForensics++ datasets with Low-Quality (light compression) and High-Quality (heavy compression). ACC(\%) and AUC(\%) are adopted, and the best performances are marked as bold. }
	\begin{tabular}{l|cc|cc}
		\toprule
		\multirow{2}{*}{Methods} & \multicolumn{2}{c|}{LQ}        & \multicolumn{2}{c}{HQ}         \\ \cline{2-5}
		& \multicolumn{1}{c}{ACC} & AUC & \multicolumn{1}{c}{ACC} & AUC  \\ \hline
		Steg. Features \cite{stegfeatures}                     & $70.97$                       & $-$   & $55.98$                        & $-$    \\
		MesoNet \cite{mesonet}                     & $70.47$                       & $-$   & $83.10$                        & $-$    \\
		Cozzolino et al. \cite{cozzo}      & $78.45$                 & $-$     &       $58.69$     & $-$\\
		Bayar et al. \cite{bayar}    & $82.97$                          & $-$        & $66.84$ & $-$\\
		Face X-ray \cite{facexray}                     & $-$                       & $61.60$  & $-$                       & $87.40$\\
		Two Branch \cite{twobranch}                   & $-$                       & $86.59$  & $-$                       & $98.70$ \\
		Xception \cite{xception}                     & $86.86$                       & $89.30$  & $95.73$                       & $96.30$   \\
		EfficientNet-B4 \cite{efficientnet}                     & $86.67$                       & $88.20$  & $96.63$                       & $99.18$\\
		Multi-attentional \cite{multiattentional}           & $86.95$                       & $87.26$  & $96.37$                       & $98.97$   \\
		$F^3$-Net \cite{f3net}                           & $90.43$                       & $93.30$  & $97.52$                       & $98.10$\\ \hline
		Ours                     & $\mathbf{95.81}$                       & $\mathbf{98.49}$  & $\mathbf{99.51}$                       & $\mathbf{99.88}$   \\
		\bottomrule
	\end{tabular}
	\label{tab:ff++}
\end{table}

\begin{table}
	\centering
	\caption{The quantitative comparisons on Celeb-DF datasets. ACC(\%) and AUC(\%) are adopted. }
	\begin{tabular}{l|cc}
		\toprule
		Methods      & ACC   & AUC   \\ \hline
		MesoNet \cite{mesonet} & $-$ & $53.6$ \\
		I3D \cite{I3D}    & $76.08$ & $83.00$    \\
		C3D  \cite{C3D}      & $78.67$ & $84.00$    \\
		FaceNetLSTM	\cite{FaceNetLSTM} & $79.83$ &    $-$   \\
		Hu et al. \cite{social}   & $80.74$ & $87.00$    \\
		Xception \cite{xception} & $89.55$ & $89.91$ \\
		FakeCatcher \cite{fakecatcher} & $91.50$ &     $-$  \\
		XcepTemporal \cite{xceptemporal}    & $97.83$ &   $-$    \\ \hline
		Ours         & $\mathbf{99.13}$ & $\mathbf{99.87}$ \\ \bottomrule
	\end{tabular}
	\label{tab:celebdf}
\end{table}

% \usepackage{multirow}
\begin{table*}
	\centering
	\caption{The quantitative comparison (ACC (\%) and AUC (\%)) on FaceForensics++ with four different manipulation methods, i.e., DeepFakes(DF) \cite{deepfake}, Face2Face(F2F) \cite{face2face}, FaceSwap(FS) \cite{faceswap}, NeuralTextures(NT) \cite{neuraltextures}. The proposed method is capable of dealing with different manipulation methods.
	}
	\setlength{\tabcolsep}{4mm}{
		\begin{tabular}{l|cc|cc|cc|cc}
			\toprule
			\multicolumn{1}{l|}{\multirow{2}{*}{Methods}} & \multicolumn{2}{c|}{DF \cite{deepfake}} & \multicolumn{2}{c|}{F2F \cite{face2face}} & \multicolumn{2}{c|}{FS \cite{faceswap}} & \multicolumn{2}{c}{NT \cite{neuraltextures}} \\ \cline{2-9}
			\multicolumn{1}{c|}{}                         & ACC        & AUC       & ACC         & AUC        & ACC        & AUC       & ACC        & AUC        \\ \hline
			Steg. Features  \cite{stegfeatures}                              & $73.64$      &    $-$   & $73.72$   &$-$      & $68.93$   & $-$       & $63.33$      &  $-$         \\
			Cozzolino et al.    \cite{cozzo}                          & $85.45 $     &    $-$   & $67.88 $      &$-$    & $73.79$   &  $-$    & $78.00$   &  $-$          \\
			Rahmouni et al.     \cite{rahmouni}                          & $85.45 $     &    $-$   & $64.23 $      &$- $  & $56.31$      &  $-$   & $60.07$      &   $-$         \\
			Bayar et al.      \cite{bayar}                          & $84.55 $     &    $-$   & $73.72 $      &$ -$    & $82.52$   & $-$          & $70.67$   &  $-$          \\
			C3D \cite{C3D}   										& $85.10$ &  $91.00$  &  $73.12$  &  $88.00$ &  $72.11$ &  $87.00$   &  $60.30$  & $59.00 $  \\
			Hu et al.          \cite{social}                          & $94.64 $     & $98.00 $    & $86.48 $      & $94.00$      & $85.27$ & $94.00$  & $80.05$      & $90.00$      \\
			MesoNet        \cite{mesonet}                               & $87.27 $     & $-$    &$ 56.20 $      &$-$  & $61.17$      & $-$     & $40.67$      &   $-$         \\
			Xception     \cite{xception}                              & $95.15 $     & $99.08$    & $83.48$     & $93.77$ & $92.09$  & $97.42$     & $77.89$      & $84.23$      \\
			Spatial-phase   \cite{spatialphase}                              & $93.48 $     & $98.50$    & $86.02$   & $94.62$ & $92.26$   & $98.10$  & $76.78$      & $80.49$      \\
			FakeCatcher    \cite{fakecatcher}                               & $94.87 $     & $-$    & $96.00$       & $-$      & $95.75$ & $-$     & $89.12$   & $-$           \\ \hline
			Ours                                &$\mathbf{99.47} $  & $\mathbf{99.79}$  & $\mathbf{99.98}$   & $\mathbf{100.00}$  & $\mathbf{98.27}$  & $\mathbf{99.46}$   & $\mathbf{93.25}$ & $\mathbf{98.61}$      \\ \bottomrule
		\end{tabular}
		\label{tab:fourmani}
	}
\end{table*}

Xception \cite{xception} is the backbone we adopted which has 12 main blocks and some feature extraction layers at the beginning.
In our experiments, the learning rate is set as $0.0003$, which is determined by experiments.
The networks are optimized by SGD with momentum=$0.9$.
The quantity of attention maps is set by experiments, and the default number is $4$, more discussion is given in \ref{sec:quantitymaps}.
%The spatial attention maps are working with shallow features generated by the first several layers and the temporal attention maps are working with the feature maps generated by the 2-th block of the backbone.
% The layer in which spatial attention maps are fed is the 1-th  block

The face forgery detection is a binary classification task, that is, gives a judgement of the tested video whether it is fake or real.
Two evaluation metrics are adopted in our experiments, Accurency rate (ACC) is the most intuitive evaluation metric.
AUC is another metric we adopted.

\begin{figure}
	\centering
	{\includegraphics[width=0.41\textwidth]{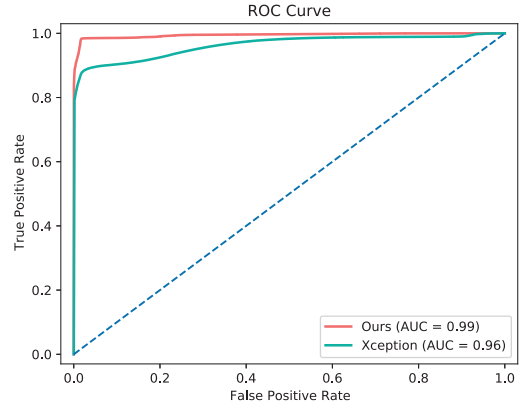}}
	\caption{ROC curves for the Xception and spatial-temporal model on HQ of FaceForensics++. }
	\label{fig:AUC}
\end{figure}

\subsection{Comparison experiments}

Comparisons are conducted among current state-of-the-art deepfake detection methods and the proposed method.
We first perform our experiments on FaceForensics++ \cite{ff++}, which has been widely tested in this field.
As aforementioned, there are two compression versions in FaceForensics++ \cite{ff++}, HQ represents the low-level compression version, and LQ represents the high-level compression version.
The performances demonstrated in Table \ref{tab:ff++} are tested on both HQ (c23) and LQ (c40) versions with ACC and AUC metrics.
The experimental results indicate that the proposed method achieves state-of-the-art performance on both versions of FaceForensics++ \cite{ff++}.
In general, the performance of most methods in the high compressed video is not as good as that in low compressed video.
This is because the video will lose a lot of texture details after high compression, which is one of the main pieces of information that networks need to pay attention to.
Since the proposed method takes into account the inconsistencies of multiple frames in the time domain, compared with the other methods, the performance degradation is relatively small.
Another noteworthy point is that, compared with the backbone Xception \cite{xception}, the proposed method has a significant improvement as shown in Fig. \ref{fig:AUC}, which is benefiting from the spatial-temporal guidance.
The Celeb-DF \cite{celebDF} datasets is also adopted. As shown in Table \ref{tab:celebdf}, although the Celeb-DF is very realistic, the proposed model can effectively capture the defects and achieve better performance than the other methods.

To evaluate the spatial-temporal model's ability to capture defects introduced by different manipulation methods, the model is trained and tested on different manipulation methods in FaceForensics++ \cite{ff++}.
As the results shown in Table \ref{tab:fourmani}, for different manipulation methods, our method achieves better performances than the other methods.
It confirms  that the proposed spatial-temporal model is capable of capturing various kinds of defects introduced by different manipulation methods.
This may be because the defects introduced by these operation methods have some common characteristics, and the attention mechanism helps to excavate these characteristics.

\begin{table}
        \centering
        \caption{Cross-dataset evaluation with AUC(\%). Trained on HQ and LQ of FaceForensics++ and tested on Celeb-DF. Our method outperforms most deepfake detection methods.}
        \begin{tabular}{l|c|cc}
        \toprule
        Methods       & FF++   & Celeb-DF   &  \\ \cline{1-3}
        Two-stream \cite{twostream}   &  $70.10$  & $53.80$  &  \\
        MesoNet   \cite{mesonet}      &  $84.70$  & $54.80$  &  \\
        FWA    \cite{fwa}        &  $80.10$  & $56.90$  &  \\
        Xception   \cite{xception}   & $ 99.70$  & $48.20 $ &  \\
        Multi-task \cite{multitask}   &  $76.30$  & $54.30$  &  \\
        Capsule  \cite{capsule}    & $ 96.60$  & $57.50$  &  \\
        DSP-FWA  \cite{fwa}     &$ 93.00$   & $64.60$  &  \\
        Two Branch \cite{twobranch}   & $ 93.18$  & $\mathbf{73.41}$ &  \\
        $F^3$-Net     \cite{f3net}   &  $98.10$  & $65.17$ &  \\
        Multi-attentional \cite{multiattentional} &$99.80$ & $67.44$ &  \\ \cline{1-3}
        Ours          &  $\mathbf{99.97}$  & $70.33$ &  \\
        \bottomrule
        \end{tabular}
        \label{tab:cross-dataset}
\end{table}

\subsection{Cross-dataset performance}

In this part, the transferability of our framework is evaluated.
The cross-dataset result is shown in Table \ref{tab:cross-dataset}.
To compare with other methods, we train our model on both HQ and LQ of FaceForensics++ \cite{ff++} and tested on Celeb-DF \cite{celebDF}.
%Results of some other methods are cited directly from their original papers and \cite{twobranch} for a fair comparison.
Since there are many differences between datasets, such as different video compression method, common scenes, camera angles, and so on, it is a challenging task for most detection methods.
All the methods have different degrees of decline in the cross-dataset task.
As the experimental results show, although our method is not specially designed for cross-dataset performance, it still has better performance than most methods.
Two-Branch \cite{twobranch} is elaborately designed for transferability and achieves better results. However, our in-dataset performance is better than theirs.
The comparison with the backbone network also confirms that our spatial-temporal model can effectively emphasize local regions, thus improving the transferability.

\begin{figure}
	\centering
	\subfigure[Capturing local defects]{\includegraphics[width=0.5\textwidth]{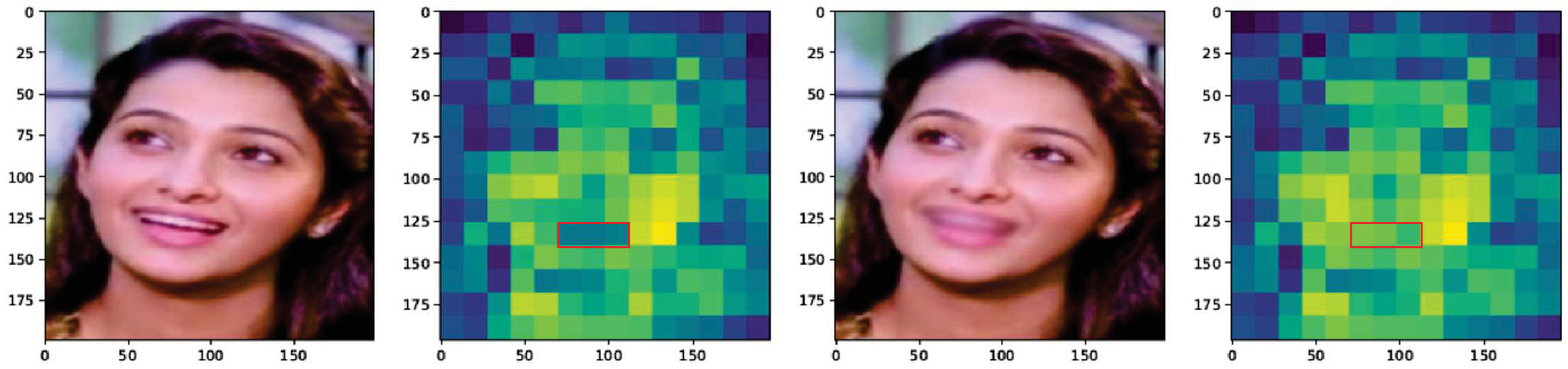}}
	\subfigure[Capturing mismatched eyes]{\includegraphics[width=0.5\textwidth]{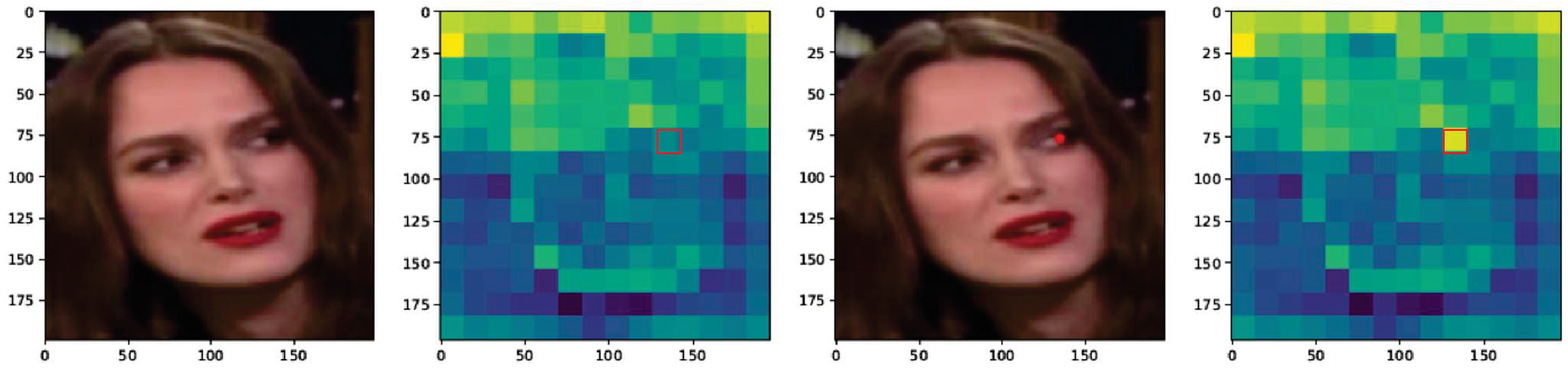}}
	\subfigure[Capturing abnormal structures]{\includegraphics[width=0.5\textwidth]{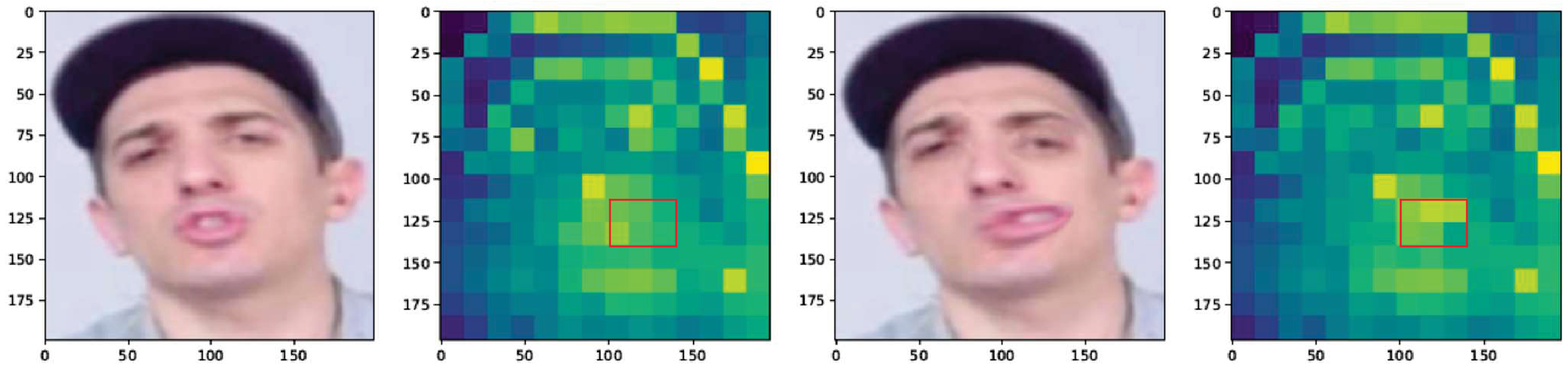}}
	\caption{ The attention maps effectively emphasize the tampered region. The highlighted area in the attention map is highly coincident with the tampered area of the real face.
	}
	\label{fig:capturingdefects}
\end{figure}

\subsection{Ability of capturing defects}

Since the defects in deepfake videos are subtle,
it may not be initiative for human eyes to discriminate the differences between the attention maps generated by real and fake faces.
To intuitively understand how the long distance attention works, we manually add the defects that we mentioned in Fig. \ref{fig:fakesample} to the frames of authentic videos, and examine the differences between attention maps of the real and tampered faces.
Although deepfake videos generally do not produce such obvious traces of forgery, the use of these obvious tampered faces helps to intuitively understand how the long distance attention can capture these local and global defects.
As shown in Fig. \ref{fig:capturingdefects}, the first column consists of real faces,
the second column consists of attention maps of the real faces, the third column is the tampered version of the first column by a certain manipulation,
and the last column consists of attention maps of the tampered faces.
The areas highlighted in the fourth column but not highlighted in the second column are marked with red boxes,
and it can be seen that they are highly coincident with the tampering position in the face.
The first row is an example of local defects.
The tampered image is Gaussian blurred to simulate the texture defects in deepfake videos.
The mouth area of the real face is blurred, and it can be seen that the attention map generated by the tampered face highlights the corresponding area.
The second row is an example of mismatched eyes. The face's right eye pupil is painted red.
Obviously, the area of the attention map corresponding to the abnormal eye is highlighted.
The last row is an example of abnormal face structures.
The mouth of the face is distorted.
Therefore, the generated attention map highlights the corresponding area of this abnormal structure.
These results indicate that the long distance attention mechanism can capture the defects in local and global perspectives.
Thus, the long distance attention mechanism is useful to generate guidance from the local and global perspectives, and make the backbone network focus on the pivotal regions.

\section{Ablation analysis}
\label{sec:ablationanalysis}

In this section, we discuss the effectiveness of the temporal attention model and the spatial attention model respectively, and further discuss the influence of model parameters on performance.

\begin{figure}
        \centering
             \subfigure[real1]{\includegraphics[width=0.11\textwidth]{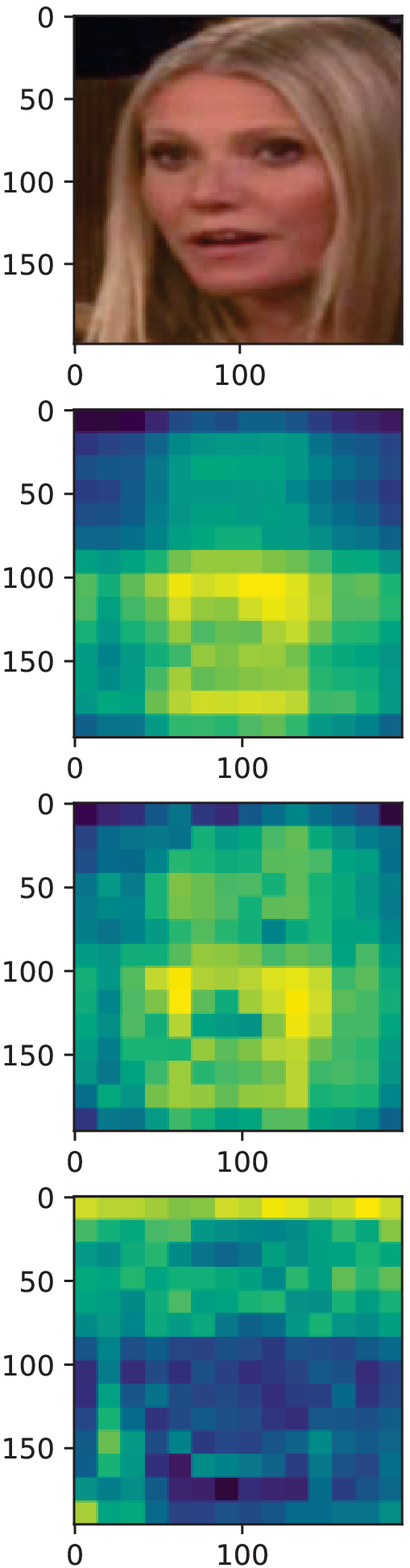}}
             \subfigure[real2]{\includegraphics[width=0.11\textwidth]{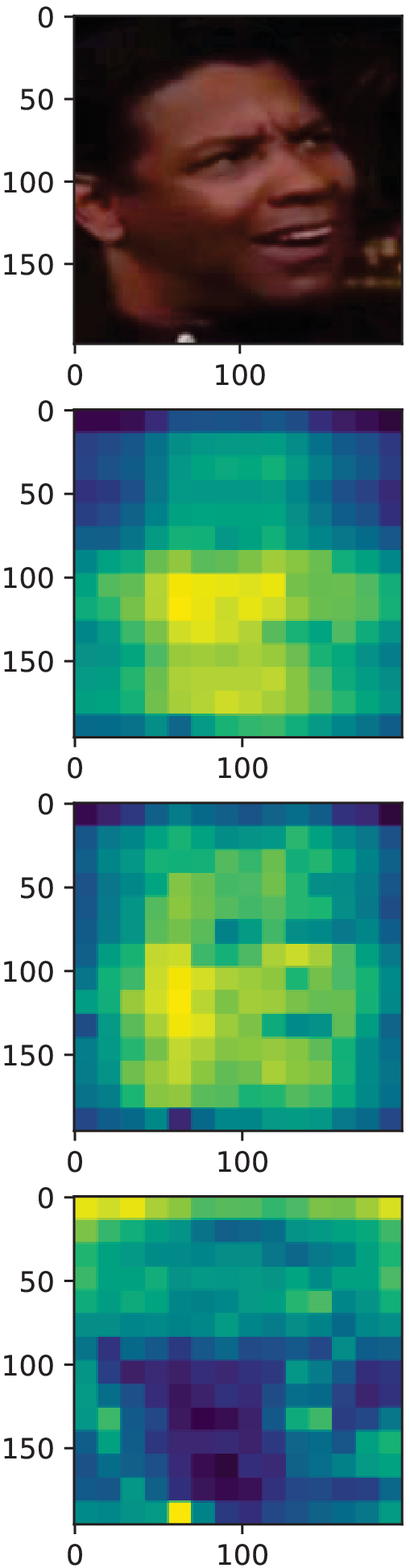}}
             \subfigure[fake1]{\includegraphics[width=0.11\textwidth]{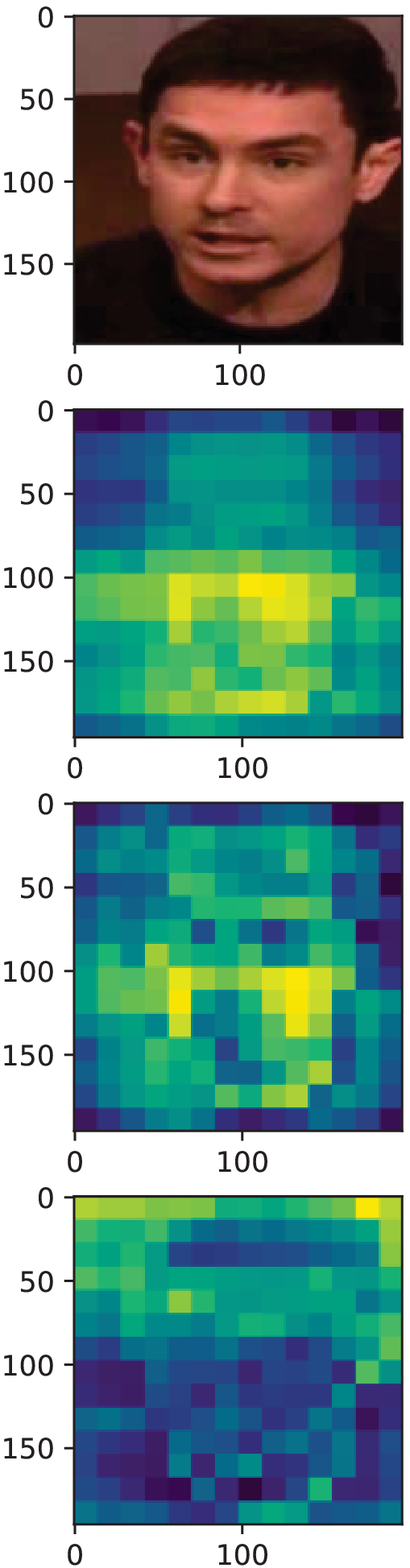}}
             \subfigure[fake2]{\includegraphics[width=0.11\textwidth]{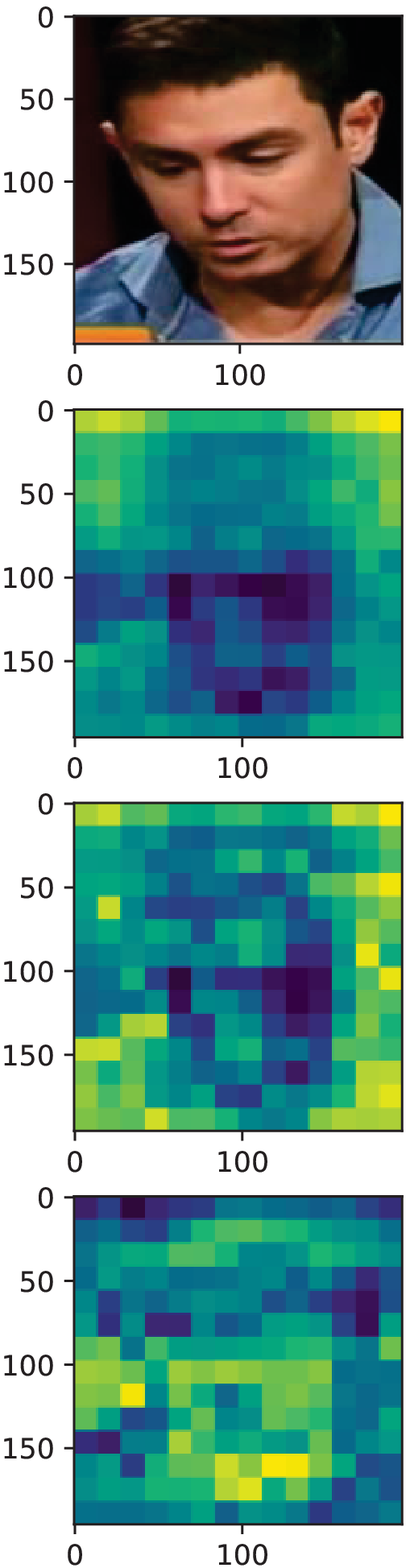}}
           \caption{Spatial attention maps generated by real videos and fake videos, brighter areas will be emphasized. }
        \label{fig:spatial_attention}
\end{figure}

\begin{figure}
        \centering
             \subfigure[real1]{\includegraphics[width=0.11\textwidth]{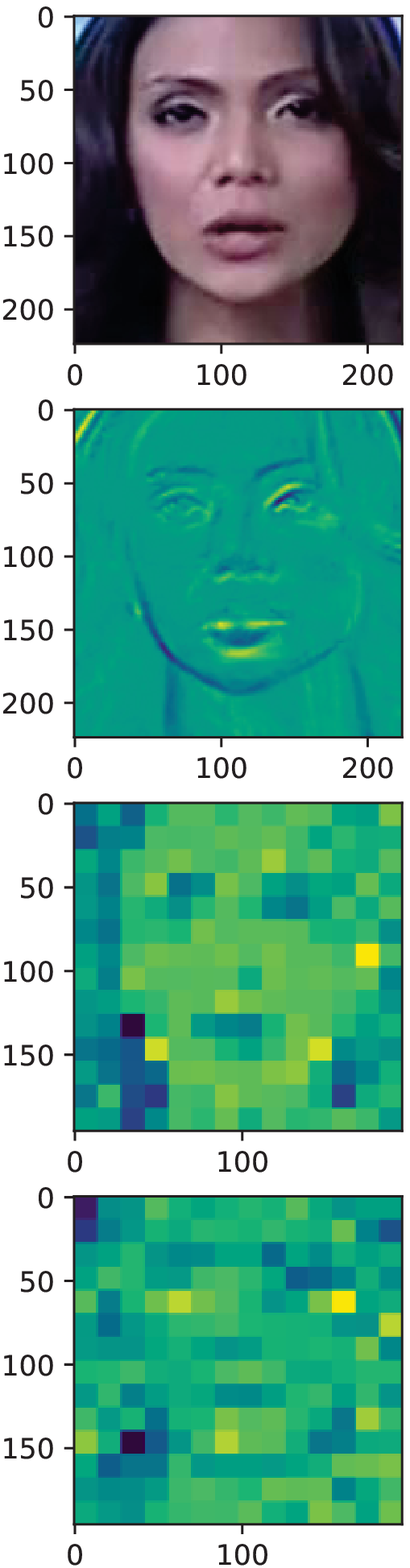}}
             \subfigure[real2]{\includegraphics[width=0.11\textwidth]{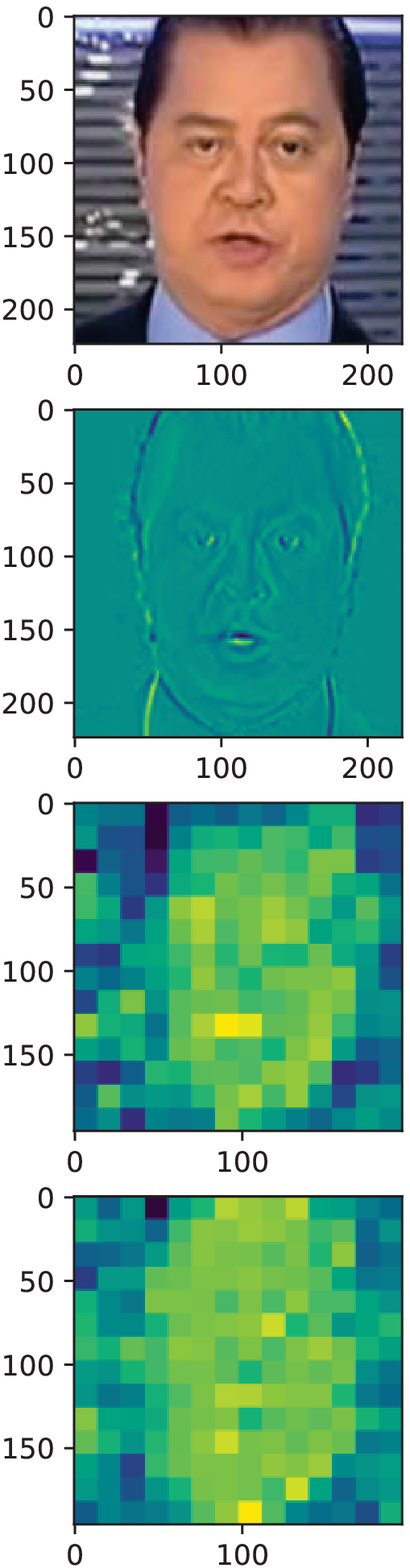}}
             \subfigure[fake1]{\includegraphics[width=0.11\textwidth]{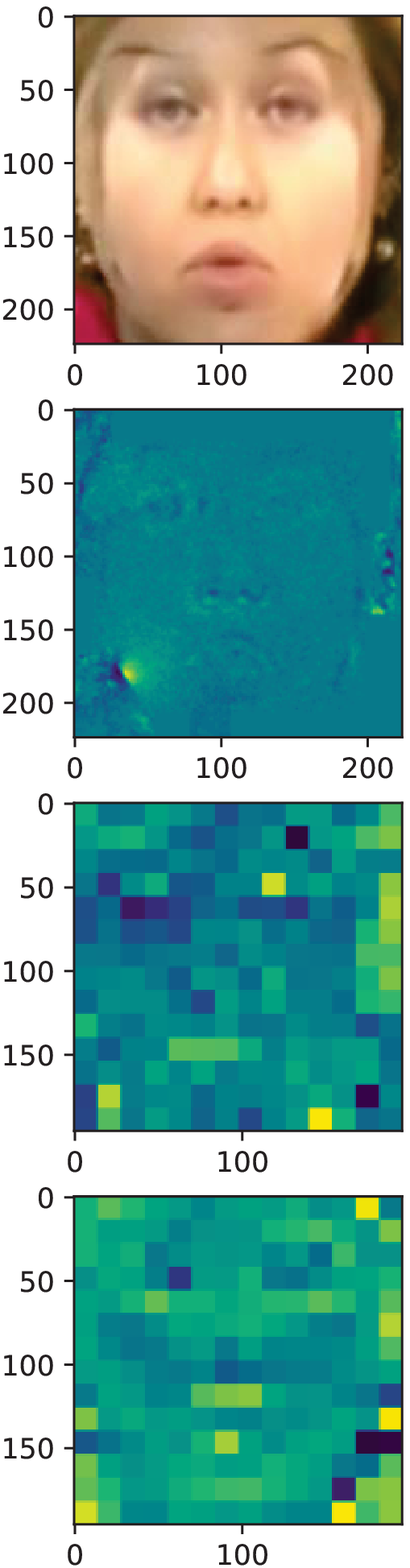}}
             \subfigure[fake2]{\includegraphics[width=0.11\textwidth]{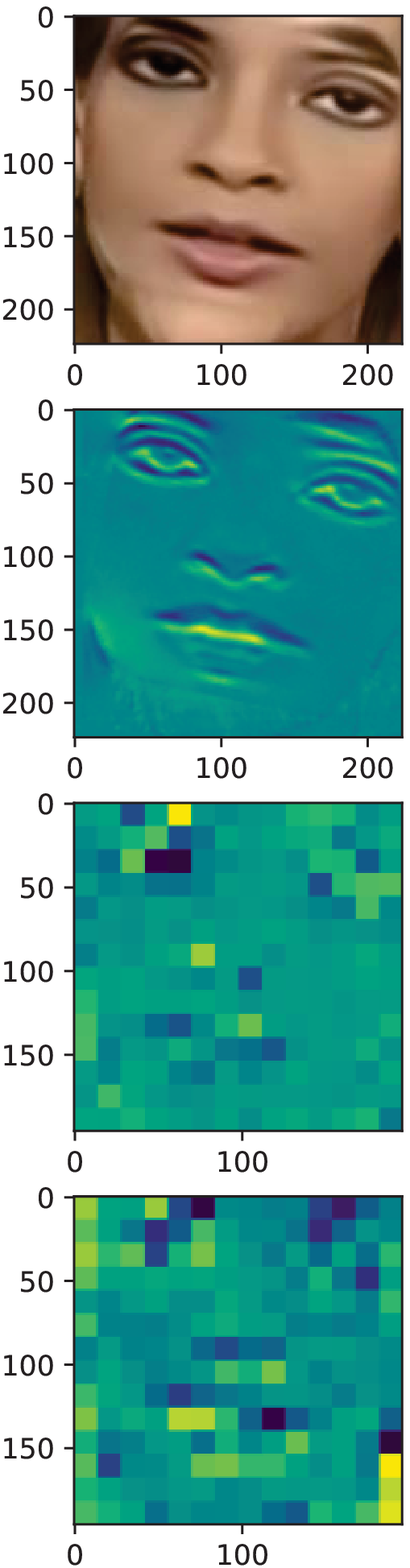}}
           \caption{Temporal attention maps generated by real videos and fake videos, brighter areas will be emphasized. }
        \label{fig:motion_attention}
\end{figure}

\subsection{Effectiveness of spatial-temporal model}

% \begin{table}
%         \centering
%         \caption{The models are trained and tested on FaceForensics++ HQ, both spatial model and temporal model are effective and have a significant improvement, the best performance is achieved by the spatial-temporal model}
%         \begin{tabular}{l|c|c}
%         \toprule
%         Model            & ACC   & \multicolumn{1}{l}{AUC} \\ \hline
%         xception         & $95.73$ & $96.30$                     \\
%         spatial          & $99.11$ & $99.76$                    \\
%         temporal         & $98.80$ & $99.85$                    \\ \hline
%         spatial-temporal & $\mathbf{99.51}$ & $\mathbf{99.88}$                    \\
%         \bottomrule
%         \end{tabular}
%         \label{tab:different models}
% \end{table}

\begin{table}
	\centering
	\caption{The models are trained and tested on FaceForensics++ HQ, both spatial model and temporal model are effective and have a significant improvement, the best performance is achieved by the spatial-temporal model.}
        \begin{tabular}{c|ccc|c}
        \hline
        Models & Xception \cite{xception} & spatial & temporal & spatial-temporal \\ \hline
        ACC(\%)   & $95.73$    & $99.11$   & $98.80$     & $\mathbf{99.51}$        \\ \hline
        AUC(\%)  & $96.30$     & $99.76$   & $99.85$    & $\mathbf{99.88}$            \\ \hline
        \end{tabular}
    	\label{tab:different models}
\end{table}

To evaluate the effectiveness of the spatial model and temporal model,
we separately use the spatial model, the temporal model, and the combination of the two models to compare the performance with the backbone.
All of the models are trained and tested on the FaceForensics++ \cite{ff++} with ACC and AUC metrics.
The comparison results are shown in Table \ref{tab:different models}.
It can be clearly seen that the proposed temporal model and spatial model both have significant performance improvement compared with the backbone.
The best performance is present by the spatial-temporal model, confirming  that both the spatial model and temporal model are effective.
More specifically, compared with the backbone network, each model has at least $3$ percent performance improvement in ACC and AUC metrics, and the combination of the two will have a better effect. At the same time,
it can be observed that the spatial model is slightly better than the temporal model.
We think this may be because the defects in the spatial domain are more common in deepfake videos.

In order to understand the guiding role of attention maps intuitively, the attention maps produced by the model are visualized.
The spatial attention maps are shown in Fig.\ref{fig:spatial_attention}.
The first two columns of attention maps are generated from real video frames, while the last two columns of attention maps are generated from forged video frames.
Although all attention maps successfully capture the semantic regions of the human face, the slight difference is that the highlight regions of spatial attention maps from fake videos are more concentrated.
This phenomenon is also reflected in the temporal attention maps.
As shown in Fig. \ref{fig:motion_attention}, the weight of temporal attention maps generated by real video frames is more uniform, while the temporal attention map generated by fake video focuses on a few areas.
We think it's caused by irregular, tiny jitters that often occur in deepfake videos, especially near the mouth and the edge of the face.
It is consistent with the highlight of the temporal attention maps of fake.

\subsection{Quantity of attention maps}
\label{sec:quantitymaps}

% \begin{table}
%         \centering
%         \caption{Comparison of models with different number of attention maps.}
%         \begin{tabular}{l|c|c}
%         \toprule
%         m            & ACC   & \multicolumn{1}{l}{AUC} \\ \hline
%         1         & $97.79$ & $99.73$                     \\
%         2          & $99.35$ & $99.62$                    \\
%         3         & $99.47$ & $\mathbf{99.94}$                    \\
%         4        & $\mathbf{99.51}$ & $99.88$                    \\
%         5        & $99.38$ & $99.91$ \\
%         \bottomrule
%         \end{tabular}
%         \label{tab:different maps}
% \end{table}

\begin{table}
	\centering
	\caption{Comparison of models with different number $m$ of attention maps.}
	\begin{tabular}{c|ccccc}
		\hline
		m       & 1     & 2     & 3     & 4     & 5     \\ \hline
		ACC(\%) & $97.79$ & $99.35$ & $99.47$ & $\mathbf{99.51}$ & $99.38$ \\ \hline
		AUC(\%) & $99.73$ & $99.62$ & $\mathbf{99.94}$ & $99.88$ & $99.91$ \\ \hline
	\end{tabular}
        \label{tab:different maps}
\end{table}

In order to enhance the diversity of the guidance generated by the spatial model and temporal model, and avoid generating guidance limited from a single latent space, multiple long distance attention modules are used in each model.
At the same time, in order to enhance the robustness and stability of the guidance, $1\times1$ convolution kernel is used to combine the guidance and generate $m$ final attention maps.
To verify the effectiveness of the multi-attention maps and explore the optimal quantity of attention maps, experiments are conducted on the influence of the quantity of attention maps on the performance of the model.
The models are trained on FaceForensics++ \cite{ff++} with the same hyper-parameters except for the quantity of attention maps.
As the result shown in Table \ref{tab:different maps}, since multiple attention maps provide more diversity of guidance,
the model using multiple attention maps has better performance than the model using a single attention map, and the best ACC is obtained with $m=4$, and the best AUC is obtained  with $m=3$.
When the number of maps increases to a certain number, blindly increasing the number cannot bring obvious performance improvement.

\subsection{Quantity of consecutive frames}
\label{sec:quantityofframes}

% \begin{table}
%         \centering
%         \caption{Comparison of models with different number of consecutive frames.}
%         \begin{tabular}{l|c|c}
%         \toprule
%         n            & ACC   & \multicolumn{1}{l}{AUC} \\ \hline
%         2          & $99.33$ & $99.85$                    \\
%         3         & $\mathbf{99.51}$ & $99.88$                    \\
%         4        & $99.48$ & $\mathbf{99.93}$                    \\
%         5        & $99.12$ & $99.76$                    \\
%         \bottomrule
%         \end{tabular}
%         \label{tab:different frames}
% \end{table}

\begin{table}
	\centering
	\caption{Comparison of models with different number $n$ of consecutive frames.}
	\begin{tabular}{c|cccc}
		\hline
		n       & 2     & 3     & 4     & 5     \\ \hline
		ACC(\%) & $99.33$ & $\mathbf{99.51}$ & $99.48$ & $99.12$ \\ \hline
		AUC(\%) & $99.85$ & $99.88$ & $\mathbf{99.93}$ & $99.76$ \\ \hline
	\end{tabular}
        \label{tab:different frames}
\end{table}

In the temporal model, multiple consecutive frames are used to mine the inconsistencies.
Although more consecutive frames carry more temporal information, too many sequences will make it difficult for the model to establish information association.
In order to explore how many consecutive frames can provide enough temporal information for the proposed model, the temporal models with different numbers of consecutive frames are used to explore the optimal number of consecutive frames.
The experimental results of different numbers of consecutive frames are shown in Table \ref{tab:different frames}.
It can be seen that 3 consecutive frames are enough for the proposed temporal model to build the information association of the patches in the time domain.

\section{Conclusion}
\label{sec:conclusion}

In this paper, we detect deepfake video from the perspective of fine-grained classification since the difference between fake and real faces is very subtle.
According to the generation defects of the deepfake generation model in  the spatial domain and the inconsistencies in the time domain,
a spatial-temporal attention model is designed to make the network focus on the pivotal local regions.
And a novel long distance attention mechanism is proposed to capture the global semantic inconsistency in deepfake.
In order to better extract the texture information and statistical information of the image, we divide the image into small patches,
and recalibrate the importance between them.
Extensive experiments have been performed to demonstrate that our method achieves state-of-the-art performance,
showing that the proposed long distance attention mechanism is capable of generating guidance from a global perspective.
Apart from the spatial-temporal model and the long distance attention mechanism,
we think a main contribution of this paper is that we confirm not only focusing on pivotal areas is important, but combining global semantics is also critical.
This is a noteworthy point, which can be a strategy to improve current models.
% combining global semantics is just as important as focusing on local areas.


\end{document}